\documentclass[conference]{IEEEtran}
\IEEEoverridecommandlockouts
\usepackage{cite}
\usepackage{amsmath,amssymb,amsfonts}
\usepackage{algorithmic}

\usepackage{graphicx}
\usepackage{textcomp}
\usepackage{booktabs}
\usepackage{multirow}
\usepackage[table]{xcolor}

\usepackage{xcolor}
\usepackage[colorlinks]{hyperref}
\def\BibTeX{{\rm B\kern-.05em{\sc i\kern-.025em b}\kern-.08em
    T\kern-.1667em\lower.7ex\hbox{E}\kern-.125emX}}

\usepackage{geometry}
\usepackage{hyperref}
\usepackage{enumitem}
\usepackage{amssymb}

\newcommand{\triangleitem}{\item[\(\triangleright\)]}

\IEEEoverridecommandlockouts

\begin{document}

\title{ConcealGS: Concealing Invisible Copyright Information in 3D Gaussian Splatting}

\author{
\IEEEauthorblockN{Yifeng Yang$^{1 \dagger}$, Hengyu Liu$^{2 \dagger}$, Chenxin Li$^{2\ddagger}$, Yining Sun$^{3}$, Wuyang Li$^{2}$\\
Yifan Liu$^{2}$, Yiyang Lin$^{2}$, Yixuan Yuan$^{2 *}$, Nanyang Ye$^{1 *}$\thanks{$^{\dagger}$Equal contribution, $^{\ddagger}$Project Lead, $^{*}$Corresponding author.}}
\IEEEauthorblockA{
\textit{$^{1}$Shanghai Jiao Tong University}
\textit{$^{2}$The Chinese University of Hong Kong}
\textit{$^{3}$Johns Hopkins University}
}
}

\maketitle

\begin{abstract}

As 3D Gaussian Splatting (3D-GS) emerges as a promising technique for 3D reconstruction and novel view synthesis, offering superior rendering quality and efficiency, it becomes crucial to ensure secure transmission and copyright protection of 3D assets in anticipation of widespread distribution. While steganography has advanced significantly in common 3D media like meshes and Neural Radiance Fields (NeRF), research into steganography for 3D-GS representations remains largely unexplored. To address this gap, we propose ConcealGS, a novel 3D steganography method that embeds implicit information into the explicit 3D representation of Gaussian Splatting. By introducing a consistency strategy for the decoder and a gradient optimization approach, ConcealGS overcomes limitations of NeRF-based models, enhancing both the robustness of implicit information and the quality of 3D reconstruction. Extensive evaluations across various potential application scenarios demonstrate that ConcealGS successfully recovers implicit information with negligible impact on rendering quality, offering a groundbreaking approach for embedding invisible yet recoverable information into 3D models. This work paves the way for advanced copyright protection and secure data transmission in the evolving landscape of 3D content creation and distribution.
Code is available at \href{https://github.com/zxk1212/ConcealGS}{https://github.com/zxk1212/ConcealGS}.
\end{abstract}


\section{Introduction}
Steganography, a technique for concealing information within digital media formats \cite{kaur2016hiding, kunhoth2023video, delina2008information} such as images, audio files, videos, and text documents, has been widely used for copyright protection, content authentication, and secure information transfer.
While traditionally focused on embedding data in images or videos, the growing ubiquity of digital content has spurred demand for more sophisticated steganographic methods. As 3D representations continue to advance, we anticipate a future where sharing 3D content captured from real-world environments or modeled in virtual spaces becomes as commonplace as sharing 2D media online, opening new avenues for steganographic applications.

Advanced methods for hiding information in 3D representations primarily utilize Implicit Neural Representation (INR) \cite{jang2024waterf, liu2023hiding,li2023StegaNeRF, ong2024ipr}, which conceptualizes images and neural networks as functions \cite{park2019deepsdf, yangjie2024image,sitzmann2020implicit}. INR expresses high-dimensional information with few parameters, enabling efficient information embedding \cite{dong2024implicit} and enhancing robustness against processing operations \cite{jang2024waterf,tondi2022robust, biswal2024steganerv}. Building on this, \cite{mildenhall2021nerf} introduced Neural Radiance Field (NeRF) for realistic 3D scene renderings. StegaNeRF \cite{li2023StegaNeRF} further leverages multilayer perceptrons to simulate light propagation, reconstruct 3D scenes from multiview images, and embed implicit information in MLP weights, using U-Net \cite{ronneberger2015u,li2024u} for decoding. However, these methods are limited by their reliance on NeRF, tying the quality of rendered images and hidden information to NeRF's capabilities.

\begin{figure}[ht]
    \centering
    \includegraphics[width=1\linewidth]{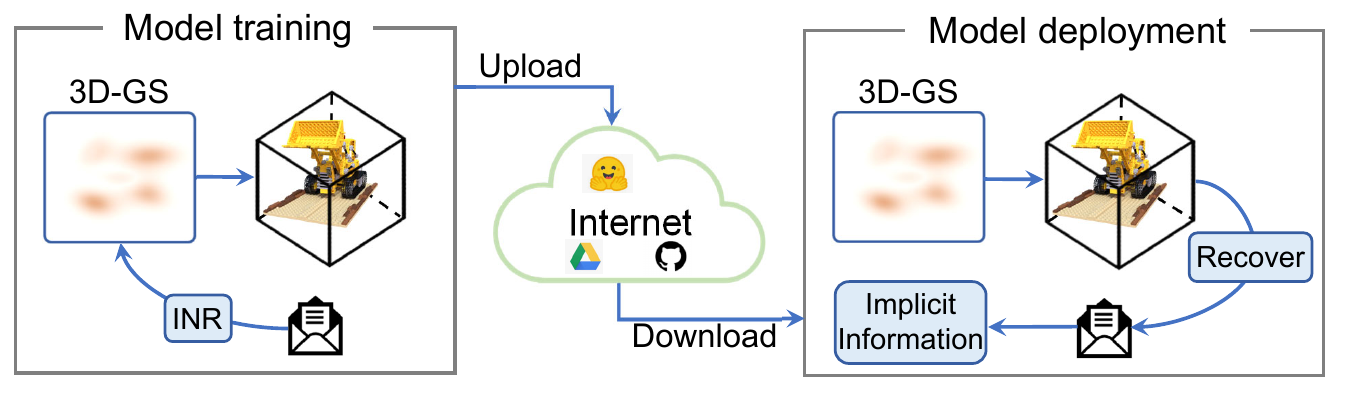}
    \caption{
llustration of the proposed copyright protection process using 3D-GS. Left: During model training, the owner embeds implicit information into the 3D-GS parameters and uploads the model. Right: During model deployment, when the shared model is used (potentially for commercial purposes), the owner can recover the embedded information from rendered images to identify copyright infringement.
    } 
    \label{fig: bad}
\end{figure}


Recently, advanced methods based on 3D Gaussian Splatting (3D-GS)~\cite{kerbl20233d} have gained popularity. Unlike NeRF, 3D-GS represents point clouds in 3D space using Gaussian distributions, achieving 3D reconstruction with fewer parameters. Recent studies \cite{huang2024sc,lee2024deblurring,li2024endosparse,liu2024lgs,fan2023lightgaussian,liu2024endogaussian} have demonstrated that 3D-GS generally leads to more efficient and higher-quality 3D representations across various tasks compared to NeRF.
Given this promise, we are eager to open the exploration for the following research questions: 

\begin{itemize}[leftmargin=*]
\triangleitem \textbf{Workable Embedding:} Can we effectively embed implicit information into 3D-GS learnable parameters?
\triangleitem \textbf{Comparative Analysis:} How does implicit information embedding in 3D-GS compare to NeRF's MLP-based implicit neural representation in terms of performance and impact on the 3D model?
\triangleitem \textbf{Robustness to Retraining:} Can the embedded information be retained after re-training on additional data?
\end{itemize}


\begin{figure*}[t]
    \centering
    \includegraphics[width=0.80\textwidth]{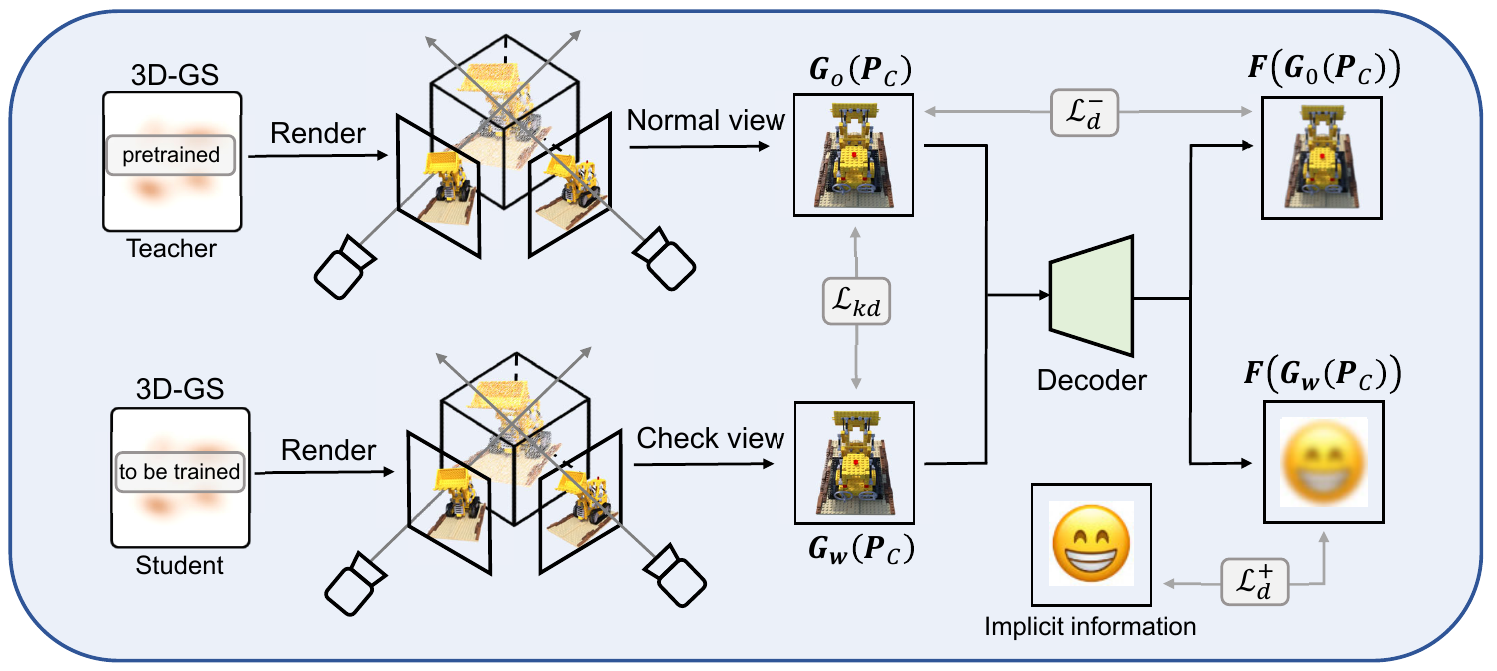}
    \caption{Overview of ConcealGS: The process involves two key stages. (1) Pre-training: A 3D-GS model is pre-trained as a teacher model for high-quality image rendering. (2) Training: Guided by the teacher model, we simultaneously train a student model and a decoder. The student model learns to render images with embedded implicit information, while maintaining visual similarity to the teacher's output. A gradient-guided optimization strategy, detailed in Section \ref{sec:2.3}, is employed to balance rendering quality and information embedding effectiveness.}
    \label{fig: attack}
\end{figure*}



To address the aforementioned questions, we consider the following scenario: As illustrated in Figure \ref{fig: bad}, during the training phase, the owner embeds implicit information into the learnable parameters of the 3D-GS and completes the 3D reconstruction. Subsequently, the reconstructed 3D representation is shared on the internet. Other users can then download and deploy this 3D representation. When the owner observes their shared 3D representation being used for commercial purposes, they can utilize a custom module to recover the embedded implicit information from the rendered images. The successful recovery of this embedded information can serve as evidence to determine whether an infringement has occurred.


Built on the superior rendering quality and rendering speed of 3D-GS~\cite{kerbl20233d}, we propose  \textbf{ConcealGS}: a 3D steganography method which embeds implicit information into the 3D explicit representation of Gaussian Splatting with almost no rendering quality decline. We use a decoder to detect the implicit information. To maintain the render quality, we distill knowledge~\cite{hinton2015distilling, li2022knowledge} from a pretrained model to ConcealGS. 
A consistency strategy is proposed for implicit information embedding. During optimization, we use a Gradient Guided Optimization strategy to balance the quality of the rendering and hidden recovery. Extensive experimental results demonstrate that the proposed method significantly outperforms NeRF-based steganography methods in both 3D reconstruction quality and the efficiency of implicit information recovery. This advancement not only enhances the potential for digital rights management in 3D content but also opens new avenues for secure information transmission within complex 3D representations.

\section{Method} 
In this section, we elaborate the proposed implicit information embedding method: ConcealGS. 
We first outline the fundamental principles of 3D Gaussian splatting in Sec.~\ref{sec:2.1}. 
In Sec.~\ref{sec:2.2}, we propose a knowledge distillation framework to embed implicit information, with a rendering consistency loss and a contrastive loss to learn effectively. 
In Sec.~\ref{sec:2.3}, we propose a gradient-guided optimization strategy which dynamically adjusts the gradient weights of the decoder layers to balance the two objectives.
The overview of the ConcealGS can be referred to in Fig.~\ref{fig: attack}.

\subsection{Preliminaries: Representation of 3D Gaussian} \label{sec:2.1}
3D-GS~\cite{kerbl20233d} employs a collection of Gaussians with various attributes to represent 3D data. Specifically, each Gaussian is defined by a covariance matrix $\boldsymbol{\Sigma}$ and a center point $\boldsymbol{X} \in \mathbb R^3$, where is the mean value of the Gaussian:
\begin{equation}
\label{formula:gaussian's formula}
\text{G}(\boldsymbol{X})=e^{-\frac{1}{2}\boldsymbol{X}^T\boldsymbol{\Sigma}^{-1}\boldsymbol{X}},
\end{equation}
where $\boldsymbol{\Sigma}$ can be decomposed into a scaling factor $\boldsymbol{s} \in \mathbb{R}^3$ and a rotation quaternion $\boldsymbol{q} \in \mathbb{R}^4$ for differentiable optimization. The rendering of 3D Gaussians entails their projection onto the image plane as 2D Gaussians~\cite{yifan2019differentiable}. The blending of ordered points for the overlap pixel can be calculated by: 
\begin{equation}
\label{eq:volume_render}
\boldsymbol{C} = \sum_{i\in N}{c_i} \alpha_i \prod_{j=1}^{i-1} (1-\alpha_i),
\end{equation}
where $\alpha \in \mathbb{R}$, $\boldsymbol{c} \in \mathbb{R}^C$ represent the opacity value and color feature, with spherical harmonics employed to capture view-dependent effects. These parameters are collectively denoted as ${\boldsymbol{G}}$, where ${\boldsymbol{G}}_i =\{\boldsymbol{X}_i, \boldsymbol{s}_i, \boldsymbol{q}_i, \alpha_i, \boldsymbol{c}_i\}$ denotes the parameters for the $i$-th Gaussian.

\subsection{Implicit Information embedding and covering} \label{sec:2.2} 
\noindent\textbf{Rendering Consistency.} Directly training a 3D representation that incorporates implicit information can substantially degrade the quality of the rendered images~\cite{li2023StegaNeRF}. 
To address this, we utilize a knowledge distillation~\cite{hinton2015distilling, li2022knowledge} strategy to maintain the rendering quality while incorporating implicit information and transferring knowledge from a pretrained model. 
In this process, a pre-trained 3D Gaussian (teacher model) guides the student model with implicit information to ensure the rendered images on known poses are not overly altered. The knowledge distillation regularization term $\mathcal{L}_{kd}$ is defined as:
 \begin{equation}
\mathcal{L}_{kd} = \left| \boldsymbol{G}_w(\boldsymbol{P}_m) - \boldsymbol{G}_o(\boldsymbol{P}_m) \right|
 \end{equation}
where $\boldsymbol{G}_w(\boldsymbol{P}_m)$ and $\boldsymbol{G}_o(\boldsymbol{P}_m)$ represent check rendered image at the camera position $\boldsymbol{P}_m$ of the student model and teacher model.
\begin{figure*}[t]
    \centering
    \includegraphics[width=0.90\textwidth]{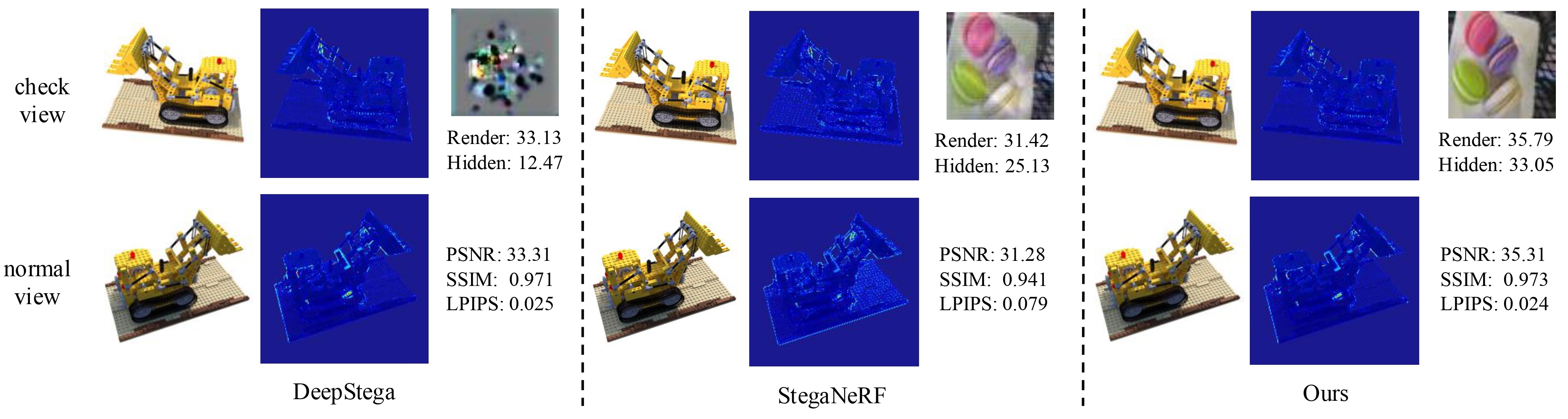}
    \caption{Qualitative comparison on NeRF-Synthetic~\cite{mildenhall2021nerf}. 
Within each column, we show the rendering images on check view and normal view, the recovered hidden image and the residual error compared to the ground truth image. We present the PSNR for scene renderings and recovered hidden images from the check view, along with the PSNR, SSIM, and LPIPS metrics for scene renderings from the normal view.
}
    \label{fig: res-all}
\end{figure*}

\noindent\textbf{Decoding Consistency.} 
ConcealGS uses a universal decoder $\boldsymbol{F}$ to decode implicit information from check view $\boldsymbol{P_c}$ rendering and original image from normal view $\boldsymbol{P}_{n} \in \{\boldsymbol{P}_{nj}\}_{j=1}^{M}$~\cite{pan2024learning}. 
To minimize the rendering quality decline on normal view, the decoder is trained as an identity network on normal view $\boldsymbol{P}_n$:
\begin{equation}
\mathcal{L}_{d}^{-} = \left| \boldsymbol{F}(\boldsymbol{G}_o(\boldsymbol{P}_n)) - \boldsymbol{G}_o(\boldsymbol{P}_n) \right| 
\label{eq2}
\end{equation}
$\mathcal{L}_{d}^{-}$ ensures that the difference between the output of the decoder and normal view remains very small. This makes the decoder's input and output as consistent as possible when normal view are used as input, thereby introducing only minimal changes or noise.

\noindent\textbf{Implicit Information processing.} 
ConcealGS encodes implicit information into the 3D explicit representation by supervising the output of the decoder on check view $\boldsymbol{P}_c$. Specifically, we minimize the loss between the output of the decoder on check view $\boldsymbol{P}_c$ and the ground truth of the implicit information:
\begin{equation}
\mathcal{L}_{d}^{+} = \left| \boldsymbol{F}(\boldsymbol{G}_w(\boldsymbol{P}_c)) - \boldsymbol{I} \right|
\label{eq1}
\end{equation}
where $\boldsymbol{I}$ is the target embedded implicit information and $\mathcal{L}_{d}^{+}$ measures the difference between the predicted implicit information and $\boldsymbol{I}$. Through optimization, 
implicit information can be recovered from rendered image on check view, which indicates the implicit information is embedded in the 3D explicit representation.  

\subsection{Gradient Guided Optimization} \label{sec:2.3}

During optimization, the process of embedding implicit information and strategy of decoding consistency will make he universal decoder trapped in the trade-off of decoding implicit information on check view ($\mathcal{L}_{d}^{+}$ in Eq \ref{eq1}) and maintaining the consistency of the normal view($\mathcal{L}_{d}^{-}$ in Eq \ref{eq2}).
To ensure that decoder effectively extracts implicit information while preserving the integrity of the rendering image and minimizing gradient conflicts, a gradient-guided optimization strategy is utilized by ConcelGS. This strategy dynamically enhances or suppresses parameter updates in specific layers by measuring the similarity between the gradients of positive and negative contrastive loss functions.

For decoder $\boldsymbol{F}$ with $N$ layers: $\{f_i\}_{i=1}^{N}$, the gradients of each layer $f_i$ with respect to $\mathcal{L}_{d}^{+}$, $\mathcal{L}_{d}^{-}$ are denoted by $\nabla_{f_i} \mathcal{L}_{d}^{+}$ and $\nabla_{f_i} \mathcal{L}_{d}^{-}$, respectively. 
We use the cosine similarity between $\nabla_{f_i} \mathcal{L}_{d}^{+}$ and $\nabla_{f_i} \mathcal{L}_{d}^{-}$, to calculate the emphasis of the current parameters on the two mentioned tasks:
\begin{equation}
    s_i=\frac{\left\langle\nabla_{f_i} \mathcal{L}_{d}^{+}, \nabla_{f_i} \mathcal{L}_{d}^{-}\right\rangle}{\left\|\nabla_{f_i} \mathcal{L}_{d}^{+}\right\|\left\|\nabla_{f_i} \mathcal{L}_{d}^{-}\right\|},
    w_i=\sigma\left(s_i\right)
\end{equation}
$s_i$ is passed through the sigmoid function $\sigma$ to obtain the weight for each layer’s gradient. During optimization, the gradient of each layer $\frac{\partial \mathcal{L}}{\partial f_i}$ is multiplied by the corresponding gradient weight $w_i$ before updating:
\begin{equation}
\frac{\partial \mathcal{L}}{\partial f_i} \leftarrow w_i \cdot \frac{\partial \mathcal{L}}{\partial f_i}, \quad \forall i \in\{1,2, \ldots, N\}
\end{equation}
Specifically, the layer with similar gradients indicates that it processes both tasks in a relatively consistent direction, meaning no significant adjustment is needed. If the gradients are dissimilar, it suggests a large discrepancy in how the layer handles the tasks, requiring suppression of the update. This dynamic adjustment mechanism coordinates the conflicting losses across different layers of the decoder, thus achieving the dual goals of implicit information extraction and the original rendering image preservation.

\begin{table}[htbp]
  \centering
  \caption{Quantitative results of rendering performance and hidden information recovery on NeRF-Snthetic~\cite{mildenhall2021nerf}.}
    \begin{tabular}{c|ccc|cc}
    \toprule
    \multirow{2}[4]{*}{Methods} & \multicolumn{3}{c|}{Scene Rendering} & \multicolumn{2}{c}{Hidden Recovery } \\
\cmidrule{2-6}          & PSNR↑ & SSIM↑ & LPIPS↓ & PSNR↑ & SSIM↑ \\
    \midrule
    3D-GS~\cite{kerbl20233d} & 34.45  & 0.9732  & 0.0220  & -     & - \\
    \midrule
    LSB~\cite{chang2003finding}   & \cellcolor[rgb]{ .992,  .973,  .706}33.05  & \cellcolor[rgb]{ .957,  .808,  .616}0.9704  & \cellcolor[rgb]{ .992,  .973,  .706}0.0253  & 6.13  & 0.1218 \\
    DeepStega~\cite{baluja2017hiding} & \cellcolor[rgb]{ .957,  .808,  .616}33.06  & \cellcolor[rgb]{ .992,  .973,  .706}0.9702  & \cellcolor[rgb]{ .957,  .808,  .616}0.0253  & \cellcolor[rgb]{ .992,  .973,  .706}11.13  & \cellcolor[rgb]{ .992,  .973,  .706}0.2265 \\
    StegaNeRF~\cite{li2023StegaNeRF} & 30.33  & 0.9381  & 0.0787  & \cellcolor[rgb]{ .957,  .808,  .616}24.60  & \cellcolor[rgb]{ .957,  .808,  .616}0.9026 \\
    ours  & \cellcolor[rgb]{ .929,  .62,  .604}33.14  & \cellcolor[rgb]{ .929,  .62,  .604}0.9717  & \cellcolor[rgb]{ .929,  .62,  .604}0.0240  & \cellcolor[rgb]{ .929,  .62,  .604}33.98  & \cellcolor[rgb]{ .929,  .62,  .604}0.9942 \\
    \bottomrule
    \end{tabular}%
  \label{tab:nerf-syn}%
\end{table}%

\begin{table}[htbp]
  \centering
  \caption{Quantitative results of rendering performance and hidden information recovery on T\&T~\cite{kerbl20233d}.}
    \begin{tabular}{c|ccc|cc}
    \toprule
    \multirow{2}[4]{*}{Methods} & \multicolumn{3}{c|}{Scene Rendering} & \multicolumn{2}{c}{Hidden Recovery } \\
\cmidrule{2-6}          & PSNR↑ & SSIM↑ & LPIPS↓ & PSNR↑ & SSIM↑ \\
    \midrule
    3D-GS~\cite{kerbl20233d} & 27.43  & 0.9190  & 0.0779  & -     & - \\
    \midrule
    LSB~\cite{chang2003finding}   & \cellcolor[rgb]{ .992,  .973,  .706}26.11  & \cellcolor[rgb]{ .992,  .973,  .706}0.8994  & \cellcolor[rgb]{ .992,  .973,  .706}0.1048  & 5.98  & 0.0826  \\
    DeepStega~\cite{baluja2017hiding} & \cellcolor[rgb]{ .957,  .808,  .616}26.17  & \cellcolor[rgb]{ .957,  .808,  .616}0.9000  & \cellcolor[rgb]{ .957,  .808,  .616}0.1047  & \cellcolor[rgb]{ .992,  .973,  .706}6.27  & \cellcolor[rgb]{ .992,  .973,  .706}0.1532  \\
    StegaNeRF~\cite{li2023StegaNeRF} & 23.48  & 0.7462  & 0.6067  & \cellcolor[rgb]{ .957,  .808,  .616}32.33  & \cellcolor[rgb]{ .957,  .808,  .616}0.9918  \\
    ours  & \cellcolor[rgb]{ .929,  .62,  .604}27.26  & \cellcolor[rgb]{ .929,  .62,  .604}0.9177  & \cellcolor[rgb]{ .929,  .62,  .604}0.0796  & \cellcolor[rgb]{ .929,  .62,  .604}37.53  & \cellcolor[rgb]{ .929,  .62,  .604}0.9935  \\
    \bottomrule
    \end{tabular}%
  \label{tab:t-t}%
\end{table}%

\section{Experiments}
\label{sec:exp}

\subsection{Experimental Settings}
\noindent\textbf{Datasets and Metrics.}\ 
We utilize two widely recognized datasets: NeRF-Synthetic~\cite{mildenhall2021nerf} and T\&T as used in~\cite{kerbl20233d}. We select four $360^{\circ}$ scenes {\textit{\{lego, drums, chair, ship\}}} from NeRF-Synthetic. The T\&T dataset includes four scenes: {\textit{\{playroom, drjohnson, truck, train\}}}. We adopt the same data splitting as described in~\cite{kerbl20233d}. Following~\cite{li2023StegaNeRF, li2024gaussianstego}, we evaluate the quality of the hidden information recovered by the decoder using two metrics: PSNR and SSIM. We employ PSNR, SSIM, and LPIPS as metrics for rendering quality of 3D-GS.

\noindent\textbf{Implementation Details.}
We set the resolution of the hidden image to $[64, 64]$ and with the same data as~\cite{li2023StegaNeRF, li2024gaussianstego}. Following~\cite{li2023StegaNeRF,li2024gaussianstego}, a simple U-Net~\cite{ronneberger2015u} is utilized as the decoder for hidden
information and the learning rate of the detector is set to $1e^{-4}$. Other settings and parameters on NeRF-Synthetic and T\&T are the same as~\cite{kerbl20233d}.

\begin{figure}[ht]
    \centering
    \includegraphics[width=0.80\linewidth]{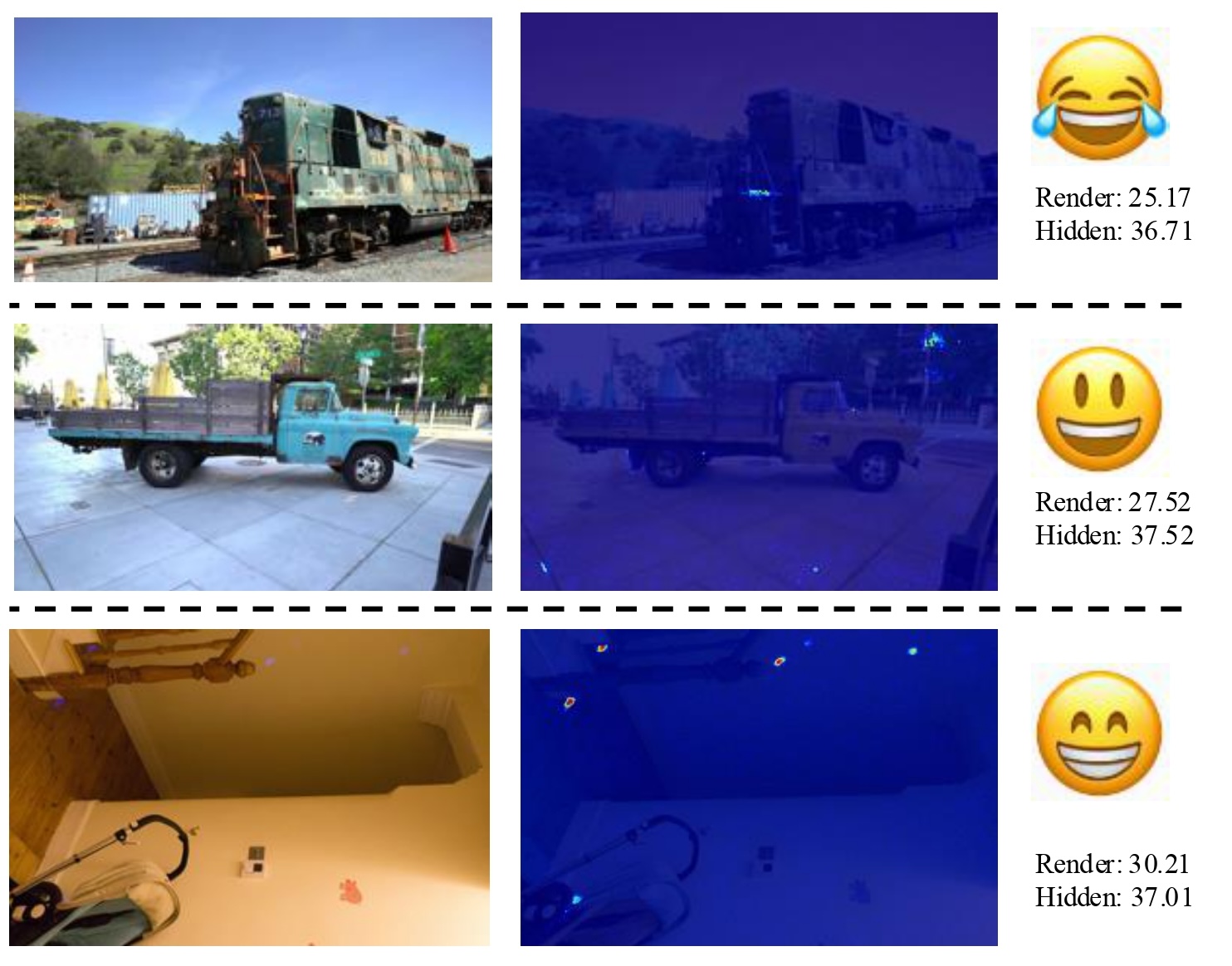}
    \caption{Results on the T\&T~\cite{kerbl20233d}. Each row displays scene rendering, residual error compared to ground truth image, and the recovered hidden image. We present the PSNR of scene rendering and hidden recovery.
} 
    \label{fig: res-ours}
\end{figure}

\subsection{Experimental Results}
We use three widely recognized techniques: LSB~\cite{chang2003finding}, DeepStega~\cite{baluja2017hiding}, and StegaNeRF~\cite{li2023StegaNeRF} for comparison. 
Experimental results on NeRF-Synthetic~\cite{mildenhall2021nerf} and T\&T~\cite{kerbl20233d} are respectively shown in Tab.~\ref{tab:nerf-syn} and Tab.~\ref{tab:t-t}. 
We observe that traditional methods, such as LSB~\cite{chang2003finding} and DeepStega~\cite{baluja2017hiding}, achieve reasonable scene rendering quality with PSNR values exceeding $26$. However, they struggle with hidden information recovery, exhibiting significantly lower PSNR across both datasets. In contrast, StegaNeRF~\cite{li2023StegaNeRF} effectively hides information in 3D representations. However, it compromises scene rendering quality, resulting in a PSNR that is $3$ lower than ours. ConcealGS demonstrates a PSNR improvement of $0.89$ and $5.20$ over other methods in scene rendering and hidden recovery. 

Compared to standard 3D-GS model (without adding any hidden information), our methods only have a $0.17$ lower PSNR and $0.0013$ lower SSIM, which indicates that ConcealGS successfully conceals implicit information in the 3D representation with almost no performance decline. Visual results are shown in Fig.~\ref{fig: res-all} and Fig.~\ref{fig: res-ours}, which also demonstrates that our approach not only provides superior rendering quality but also significantly enhances hidden information recovery compared to other methods.

\subsection{Ablation Studies}
To prove the effectiveness of each component of our method, we conduct ablation experiments on the T\&T~\cite{kerbl20233d} dataset and experimental results are presented
in Tab.~\ref{tab:ablation}, where \textit{w.o. decoder} means directly recover hidden message from check pose without the decoder, \textit{w.o. consis.} devotes our methods without the consistency strategy proposed in Sec.~\ref{sec:2.2}, and \textit{w.o. grad.} devotes the model without the gradient guided optimization proposed in Sec.~\ref{sec:2.3}. We can observe that the use of decoder can help prevent the model from overfitting on the hidden information to protect scene rendering performance. The use of the consistency strategy and the gradient guided optimization enables our method adaptively updated according to the influence of different losses during the optimization process to achieve a good trade-off between the performance of the rendering scene and the hidden recovery.

\begin{table}[htbp]
  \centering
  \caption{Ablation study of different components of ConcealGS.
Results are conducted on T\&T~\cite{kerbl20233d}.}
    \begin{tabular}{c|ccc|cc}
    \toprule
    \multirow{2}[4]{*}{Methods} & \multicolumn{3}{c|}{Scene Rendering} & \multicolumn{2}{c}{Hidden Recovery } \\
\cmidrule{2-6}          & PSNR↑ & SSIM↑ & LPIPS↓ & PSNR↑ & SSIM↑ \\
    \midrule
    Ours  & \cellcolor[rgb]{ .929,  .62,  .604}27.26  & \cellcolor[rgb]{ .929,  .62,  .604}0.9177  & \cellcolor[rgb]{ .929,  .62,  .604}0.0796  & \cellcolor[rgb]{ .957,  .808,  .616}37.53  & \cellcolor[rgb]{ .957,  .808,  .616}0.9935 \\
    \midrule
    w.o. Decoder & \cellcolor[rgb]{ .992,  .973,  .706}26.25  & \cellcolor[rgb]{ .992,  .973,  .706}0.8965  & \cellcolor[rgb]{ .992,  .973,  .706}0.0884  & \cellcolor[rgb]{ .929,  .62,  .604}37.56  & \cellcolor[rgb]{ .929,  .62,  .604}0.9948 \\
    w.o. Consis. & 25.74  & 0.8542  & 0.1265  & 37.13  & 0.9924 \\
    w.o. Grad. & \cellcolor[rgb]{ .957,  .808,  .616}26.93  & \cellcolor[rgb]{ .957,  .808,  .616}0.9043  & \cellcolor[rgb]{ .957,  .808,  .616}0.0812  & \cellcolor[rgb]{ .992,  .973,  .706}37.43  & \cellcolor[rgb]{ .992,  .973,  .706}0.9912 \\
    \midrule
    3D-GS & 27.43  & 0.9190  & 0.0779  & -     & - \\
    \bottomrule
    \end{tabular}%
  \label{tab:ablation}%
\end{table}%

\subsection{Robustness Analysis}
\begin{figure}[ht]
    \centering
    \includegraphics[width=0.95\linewidth]{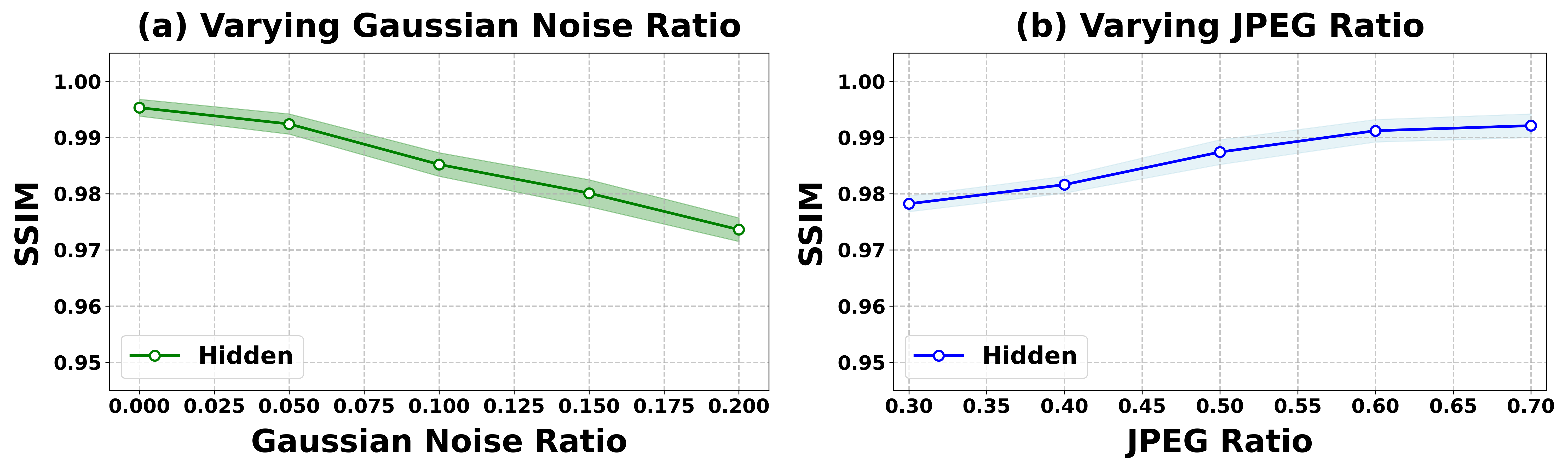}
    \caption{Robustness Analysis over various (a) Gaussian blur and (b) JPEG compression ratio. 
SSIM of hidden recovery on T\&T~\cite{kerbl20233d} is shown.} 
    \label{fig: res-rboust}
\end{figure}

The results of robustness analysis are shown in Fig.~\ref{fig: res-rboust}, where the image is scaled in $[0, 1]$ and a higher JPEG ratio means more information from the original image is retained. The SSIM values reflect mean performance across varying conditions (only $0.02$ decreased). These results suggest that the method effectively maintains the integrity of hidden information, when subjected to Gaussian Blur and JPEG compression, which indicates that our method could remain robust under the application in real-world and complex environments with various noises.

\section{Conclusion}
As 3D content increasingly becomes a core component of various applications, it is crucial to explore methods for embedding information into 3D-GS. In this paper, we propose ConcealGS, which embeds implicit information into the 3D explicit representation with almost no rendering quality decline. 
Experimental results have shown that our method outperforms other steganography methods in both rendering quality and efficiency, as well as implicit information recovery.
Extension studies demonstrate that the key components required to implement the technique are effective, and investigate the effectiveness of embedding implicit information into 3D-GS and its impact on rendering quality. 
This paper introduces a novel exploration of ownership identification within the framework of Gaussian Splatting, highlighting the need for increased attention and effort dedicated to related issues.

\bibliographystyle{IEEEbib}
\bibliography{big}

\end{document}